\documentclass[11pt,a4paper]{article}
\usepackage[hyperref]{acl2020}
\usepackage{times}
\usepackage{latexsym}

\usepackage{microtype}
\usepackage{amsmath}
\usepackage{amssymb}
\usepackage{booktabs}
\usepackage{multirow}
\usepackage{multicol}
\usepackage{graphicx}

\aclfinalcopy 

\newcommand\parheader[1]{{\bf \smallskip \noindent #1.}}
\newcommand\parheadernodot[1]{{\bf \smallskip \noindent #1}}

\title{Explicit Pairwise Word Interaction Modeling Improves Pretrained Transformers for English Semantic Similarity Tasks}

\author{Yinan Zhang, Raphael Tang, \and Jimmy Lin\vspace{0.1cm}\\
  David R. Cheriton School of Computer Science\\
  University of Waterloo
}

\date{}

\begin{document}
\maketitle
\begin{abstract}
In English semantic similarity tasks, classic word embedding-based approaches explicitly model pairwise ``interactions'' between the word representations of a sentence pair.
Transformer-based pretrained language models disregard this notion, instead modeling pairwise word interactions globally and implicitly through their self-attention mechanism.
In this paper, we hypothesize that introducing an explicit, constrained pairwise word interaction mechanism to pretrained language models improves their effectiveness on semantic similarity tasks.
We validate our hypothesis using BERT on four tasks in semantic textual similarity and answer sentence selection.
We demonstrate consistent improvements in quality by adding an explicit pairwise word interaction module to BERT.
\end{abstract}

\section{Introduction}
A substantial body of literature in the field of natural language processing is devoted to the architectural design of word embedding-based neural networks.
Over the years, painstaking progress has been made toward developing the most effective network components.
Important advancements include hierarchical attention~\cite{yang2016hierarchical}, multi-perspective convolutions~\cite{he2015multi}, and tree-structured networks~\cite{tai2015improved}.

With the rise of the transformer-based pretrained language models, however, many of these components have been all but forgotten.
Nowadays, the dominant paradigm is to pretrain a transformer~\cite{vaswani2017attention} on large text corpora, then fine-tune on a broad range of downstream single-sentence and sentence-pair tasks alike.
Prominent examples include BERT~\cite{devlin2019bert} and XLNet~\cite{yang2019xlnet}, which currently represent the state of the art across many natural language understanding tasks.

Self-evidently, these models dispense with much of the age-old wisdom that has so well guided the design of neural networks in the past.
Perhaps, that's the beauty of it all:\ a simple, universal architecture just ``works.''
However, it certainly begs the following question:\ what neural architectural design choices can we use from the past?

In this paper, we precisely explore this question in the context of semantic similarity modeling for English.
For this task, one important component is the very deep pairwise word interaction (VDPWI) module, first introduced in \citet{he2016pairwise}, which serves as a template for many succeeding works~\cite{lan2018neural}.
Conceptually, they propose to explicitly compute pairwise distance matrices for the distinct word representations of the two sentences.
The matrices are then fed into a convolutional neural network, which treats semantic similarity modeling as a pattern recognition problem.
Clearly, transformers lack such an explicit mechanism, instead modeling pairwise word interactions in an unconstrained, implicit manner through self-attention.

We take the anachronistic position that the pairwise word interaction module is still useful.
Concretely, we hypothesize that appending this module to pretrained transformers increases their effectiveness in semantic similarity modeling---we argue that this module is more than a historical artifact.
Using BERT~\cite{devlin2019bert}, a pretrained transformer-based language model, we validate our hypothesis on four tasks in semantic textual similarity and answer sentence selection.

Our core contribution is that, to the best of our knowledge, we are the first to explore whether incorporating the pairwise word interaction module improves pretrained transformers for semantic similarity modeling.
We consistently improve the effectiveness of BERT on all four tasks by adding an explicit pairwise word interaction module.


\section{Background and Related Work}

\begin{figure}
    \centering
    \includegraphics[scale=0.24]{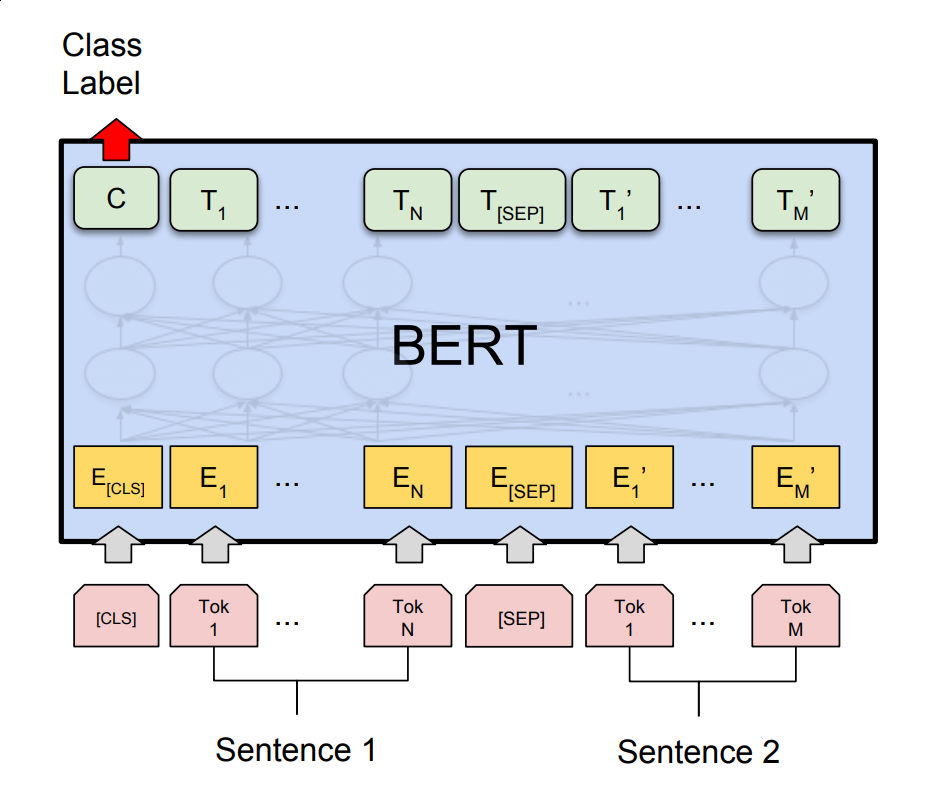}
    \caption{An illustration of BERT for sentence-pair tasks, taken from \citet{devlin2019bert}.}
    \label{fig:bert}
\end{figure}

Presently, the predominant approach to many NLP tasks is to first train an expressive language model (LM) on large text corpora, then fine-tune it on downstream task-specific data.
One of the pioneers of this approach, \citet{peters2018deep} pretrain their bidirectional long short-term memory network (BiLSTM; \citealp{hochreiter1997long}), called ELMo, on the Billion Word Corpus~\cite{chelba2014one}.
Then, for each task-specific neural network, they use the contextualized LM embeddings in place of the usual GloVe- or word2vec-based word embeddings~\cite{pennington2014glove, mikolov2013distributed}, fine-tuning the entire model end-to-end.
Using this method, they achieve state of the art across question answering, sentiment classification, and textual entailment.

\parheader{Pretrained transformers}
Recent, transformer-based pretrained language models~\cite{vaswani2017attention} disregard the task-specific neural network altogether.
Instead, the language model \textit{is} the downstream model.
\citet{devlin2019bert} are the first to espouse this approach, calling their bidirectional transformer-based model BERT.
They pretrain BERT using a cloze and next sentence prediction task on Wikipedia and BooksCorpus~\cite{zhu2015aligning}, then swap out the LM output layer with a task-specific one at fine-tuning time.

Concretely, during fine-tuning, a word-tokenized sentence pair $s_1 = \{w_{11}, \dots, w_{n1}\}$ and $s_2 = \{w_{12}, \dots, w_{m2}\}$ is first encoded as $\texttt{[CLS]} \oplus s_1 \oplus \texttt{[SEP]} \oplus s_2 \oplus \texttt{[SEP]}$, where $\oplus$ denotes concatenation, and \texttt{[CLS]} and \texttt{[SEP]} are special class and separator tokens.
Next, BERT ingests the input into a sequence of layers composed of nonlinear positionwise operators and multiheaded self-attention mechanisms, matching the transformer model---see \citet{vaswani2017attention} for specific details.
Crucially, the pairwise word interaction modeling occurs in the self-attention mechanisms, defined as
\begin{gather*}
    \text{Attn}(Q, K, V) = \text{softmax}\left(cQ K^T\right) V\\
    \text{SelfAttn}(X) = \text{Attn}(W_q X^{(l)}, W_k X^{(l)}, W_v X^{(l)})
\end{gather*}
where $c$ is a scaling constant, $W_q, W_k, W_v \in \mathbb{R}^{d\times h}$ are linear operators, and $X^{(l)} \in \mathbb{R}^{h \times L}$ is the stacked word representations at layer $l$ across an input of length $L$.
A minor point is that, for multiheaded attention, there are $\frac{h}{d}$ attention heads ($d$ divides $h$), the output representations of which are concatenated.
The key point is that this mechanism models pairwise context in a global and unconstrained manner; that is, any pair of words---even among the same sentence or the same word itself---is free to attend to each other.

Finally, for classification tasks, BERT passes the final representation of the \texttt{[CLS]} token through a softmax layer across the classes---see Figure~\ref{fig:bert}.
The entire model is fine-tuned end-to-end.

\section{Our Approach}

Given a tokenized sentence pair $s_1$ and $s_2$, \citet{he2016pairwise} first embed each word using shallow GloVe word embeddings~\cite{pennington2014glove}, pretrained on Wikipedia and GigaWord-5.
They then use BiLSTMs for modeling the context of input sentences, obtaining forward and backward context vectors $\pmb u^f_{1:|s_1|}, \pmb u^b_{1:|s_1|}$ and $\pmb v^f_{1:|s_2|}, \pmb v^b_{1:|s_2|}$ for $s_1$ and $s_2$---the superscript indicates directionality:~$f$ for forward and $b$ backward.

\parheader{Pairwise interaction layer}
From these context vectors, the distance between all context vectors across both sentences are computed to obtain a similarity cube (SimCube) of size $\mathbb{R}^{4\times k\times|s_1|\times|s_2|}$, where $k$ is the length of the similarity vector:
\begin{gather*}
\text{SimCube}[1, :, i, j] = \text{coU}(\pmb u^f_i, \pmb v^f_j)\\
\text{SimCube}[2, :, i, j] = \text{coU}(\pmb u^b_i, \pmb v^b_j)\\
\text{SimCube}[3, :, i, j] = \text{coU}(\pmb u^f_i + \pmb u^b_i, \pmb v^f_j + \pmb v^b_j)\\
\text{SimCube}[4, :, i, j] = \text{coU}(\pmb u^f_i \oplus \pmb u^b_i, \pmb v^f_j \oplus \pmb v^b_j)
\end{gather*}

\citet{he2016pairwise} define the comparison unit (coU) as coU$(\pmb u, \pmb v) = [\delta(\pmb u, \pmb v), \lVert \pmb u - \pmb v \rVert_2, \pmb u \cdot \pmb v]$, where $\delta$ denotes the cosine distance between two vectors.
The similarity cube is finally reshaped into $\mathbb{R}^{4k\times|s_1|\times|s_2|}$.
To reduce the effects of unimportant interactions, \citet{he2016pairwise} further apply a pairwise focus function and reduce their corresponding magnitudes by a factor of ten.

\parheader{Classification}
The problem is then converted to a pattern recognition one, where a 19-layer convolutional neural network models the patterns of strong pairwise interactions in the similarity cube. 
A final softmax layer is used for classification.

\subsection{BERT with VDPWI}

We use the same procedure as \citet{he2016pairwise} for word interaction modeling, except that we feed sentence input pairs to BERT~\cite{devlin2019bert} for context modeling as the first step. The contextualized embeddings from BERT are used in the downstream model for constructing similarity cube, and the entire model is fine-tuned end-to-end.

\parheader{Sentence encoding schemes}
We also explore the effectiveness of different encoding methods, as well as the contribution of the BiLSTMs in our experimental settings:

\begin{itemize}
\item \textbf{Joint vs.\ separate encoding}: we jointly or separately encode the sentence pair for BERT.
\item \textbf{Removing the BiLSTM}: we experiment with keeping or removing the BiLSTM.
\end{itemize}
In the first scheme, for joint encoding, we concatenate the tokens from the two sentences and use the regular \texttt{[SEP]} token to mark the end of the first sentence.
For separate encoding, we feed the sentences to BERT one at a time, so the two sentences do not interact with each other before the explicit interaction modeling.

In the second scheme, our motivation for removing the BiLSTM is that pretrained transformers already provide deep contextualized word embeddings, so further context modeling may be unnecessary---we may need to perform explicit pairwise word interaction modeling only.
Note that, since different forward and backward context vectors exist only with the BiLSTM, the SimCube without BiLSTMs is in $\mathbb{R}^{k \times |s_1| \times |s_2|}$.

We represent separate and joint encoding for BERT$_\text{BASE}$ by appending ``SEP'' or ``JOINT'', respectively, to the subscript of the model name.
We indicate the removal of the BiLSTM by appending ``$-$ BiLSTM'' to the name.

\section{Experimental Setup}
We run our experiments on machines with two Titan V GPUs and CUDA v10.0.
Our models are implemented in PyTorch v1.2.0.

\begin{table*}[t]
\centering 
\begin{tabular}{rlccccc}
\toprule[1pt]
\multirow{2}{*}{\#} & \multirow{2}{*}{Model} & STS-B & WikiQA & TrecQA & SICK\\ \cmidrule(l){3-6} 
& & $r/\rho$ & MAP/MRR & MAP/MRR & $r/\rho$ \\ \midrule
1 & PWIM~\cite{liu2019incorporating} & 74.4/71.8  & 70.9/72.3 & 75.9/82.2 & 87.1/80.9 \\
2 & BERT$_\text{BASE}$~\cite{devlin2019bert}  & 84.7/\underline{83.9}  & \underline{76.3}/\underline{77.6}  & 81.2/86.2  & 87.9/82.3 \\ \midrule
3 & BERT$_\text{BASE, SEP}$ $+$ PWIM  & 84.7/\underline{83.9}  & 70.5/71.6  & 69.2/72.4 & 88.0/83.6 \\ 
4 & BERT$_\text{BASE, JOINT}$ $+$ PWIM  & 84.7/\underline{83.9}  & \textbf{76.6/78.0}  & \textbf{83.7/87.9} & \underline{88.5}/\underline{83.8} \\ 
5 & BERT$_\text{BASE, SEP}$ $+$ PWIM $-$ BiLSTM  & \textbf{85.2/84.0}  & 70.6/72.0  & 68.7/72.5 & \underline{88.5}/83.7 \\ 
6 & BERT$_\text{BASE, JOINT}$ $+$ PWIM $-$ BiLSTM  & \underline{85.0}/83.7  & 73.0/74.5  & \underline{82.7}/\underline{87.5} & \textbf{88.8/84.0} \\ 
\bottomrule[1pt]
\end{tabular}
\caption{Test results on different datasets. Best results are bolded; second best underlined.}
\label{table:results}
\end{table*}
\subsection{Datasets}
We conduct experiments on two question-answering (QA) datasets and two semantic similarity datasets, all in English:

\parheadernodot{WikiQA}~\cite{yang2015wikiqa} comprises question--answer pairs from Bing query logs. We follow their preprocessing procedure to filter out questions with no correct candidate answer sentences, after which 12K binary-labeled pairs are left.

\parheadernodot{TrecQA}~\cite{wang2007jeopardy} is an open-domain QA dataset from information retrieval conferences, consisting of 56K question--answer pairs. 
 
\parheader{STS-B}
The Semantic Textual Similarity Benchmark~(STS-B; \citealp{cer2017semeval}) contains sentence pairs drawn from news headlines, video and image captions, and natural language inference data.
Human annotators assign to each pair a similarity score between one and five, inclusive.
 
\parheadernodot{SICK}~\cite{marelli2014semeval} consists of 10K sentence pairs originally from Task 1 of the SemEval 2014 competition. A similarity score between one and five, inclusive, is provided for each pair.

\smallskip \noindent SICK and STS-B are evaluated using Pearson's $r$ and Spearman's $\rho$, and TrecQA and WikiQA using mean average precision (MAP) and mean reciprocal rank (MRR).

\subsection{Training and Hyperparameters}
For fine-tuning BERT, we follow a similar procedure to \citet{devlin2019bert}.
Specifically, we perform grid search across the learning rate in $\{5, 4, 3, 2\} \times 10^{-5}$ and the number of epochs in $\{5,4,3,2\}$, choosing the configuration with the best development set scores.
Following the original setup, we use the Adam optimizer~\cite{kingma2014adam} with a batch size of 32.
For our experiments on SICK and STS-B, which use noncategorical scores, we minimize the Kullback--Leibler divergence, while we use the NLL loss on \mbox{WikiQA} and TrecQA, which are classification tasks; these objective functions are standard on these datasets~\cite{he2016pairwise}.

For training the pairwise word interaction model, following \citet{he2016pairwise}, we use the RMS\-Prop optimizer~\cite{tieleman2012lecture} with a batch size of 8.
To tune the hyperparameters on the development set, we run random search across learning rates in the interval $[5\times 10^{-5}, 5\times 10^{-4}]$ and number of epochs between 3 and 15, inclusive.

\section{Results}
We present our results in Table~\ref{table:results}.
The original VDPWI model results (first row) for WikiQA, TrecQA, and SICK are copied from \citet{liu2019incorporating}, while we train their model on STS-B, which they do not use.
The second row is the result from directly fine-tuning BERT on the four datasets.
We report our BERT with VDPWI results in rows 3--6.

\subsection{Model Quality}

For all four datasets, we find that adding explicit PWI modeling improves the effectiveness of BERT in the original joint encoding scheme---see rows 2 and 4, where we observe an average improvement of 0.9 points. The one-sided Wilcoxon signed-rank (WSR) test reveals that this difference is statistically significant ($p < 0.05$).

Although no single setting achieves the best result on all datasets---i.e., the best numbers appear in different rows in the table---two of our methods (rows 4 and 6) consistently improve upon the original BERT (row 2). Differences between BERT (row~2) and BERT with VDPWI without the BiLSTM (row 6) are not statistically significant according to the one-sided WSR test ($p > 0.05$).


\subsection{Encoding Scheme Analysis}
For joint versus separate sentence encoding schemes, we observe that, on all but STS-B, joint encoding achieves better results than separate encoding---see rows 3 and 5, which represent the separate encoding scheme, and rows 4 and 6, which represent the joint scheme.
With or without the BiLSTM, we find that separate encoding results in a degenerate solution on TrecQA, where the model underperforms the original nonpretrained model (row 1)---the gap between separate and joint encoding can be up to 14 points.
Adjusting for multiple comparisons using the Holm--Bonferroni correction, one-sided WSR tests reveal significant differences ($p < 0.05$) between all four separate--joint encoding pairs, except for the jointly encoded BERT with VDPWI (row 4) and the separately encoded BERT with the BiLSTM-removed VDPWI~(row 5; $p > 0.05$).
We conclude that, to avoid potentially degenerate solutions, jointly encoding the sentences is necessary.

For the BiLSTM ablation experiments, we do not find a detectably significant difference in keeping or removing the BiLSTM according to the two-sided WSR test ($p > 0.05$), corrected using the Holm--Bonferroni method.
Additionally, the magnitudes of the differences in the results are minor---compare rows 3 and 5, and 4 and 6.
We conclude that incorporating the BiLSTM may not be entirely necessary; the pairwise interaction layer and convolutional classifier stack suffices.

\section{Conclusions and Future Work}
We explore incorporating explicit pairwise word interaction modeling into BERT, a pretrained transformer-based language model.
We demonstrate its effectiveness on four tasks in English semantic similarity modeling.
We find consistent improvements in quality across all datasets.

One line of future work involves applying other neural network modules within and on top of pretrained language models.
Another obvious extension to this work is to examine other pretrained transformers, such as RoBERTa~\cite{liu2019roberta} and XLNet~\cite{yang2019xlnet}.

\section*{Acknowledgments}

This research was supported by the Natural Sciences and Engineering Research Council (NSERC) of Canada, and enabled by computational resources provided by Compute Ontario and Compute Canada.

\bibliographystyle{acl_natbib}

\end{document}